\documentclass[10pt, a4paper]{article}

\usepackage{lrec-coling2024} 
\usepackage{enumitem}
\usepackage{multirow}
\usepackage{makecell}

\title{Characteristic AI Agents via Large Language Models}

\name{Xi Wang\textsuperscript{1, 2}, Hongliang Dai\textsuperscript{1, 2}, Shen Gao\textsuperscript{3}, Piji Li\textsuperscript{1, 2}$^{*}$\thanks{$^{*}$ Corresponding author.}} 

\address{
\textsuperscript{\rm 1} College of Computer Science and Technology,\\
Nanjing University of Aeronautics and Astronautics, China\\
\textsuperscript{\rm 2} MIIT Key Laboratory of Pattern Analysis and Machine Intelligence, Nanjing, China\\
\textsuperscript{\rm 3} School of Computer Science and Technology, Shandong University, China\\
{\{xiwang,hldai,pjli\}@nuaa.edu.cn, shengao@sdu.edu.cn}}

\abstract{
The advancement of Large Language Models (LLMs) has led to significant enhancements in the performance of chatbot systems. Many researchers have dedicated their efforts to the development of bringing characteristics to chatbots. While there have been commercial products for developing role-driven chatbots using LLMs, it is worth noting that academic research in this area remains relatively scarce. Our research focuses on investigating the performance of LLMs in constructing Characteristic AI Agents by simulating real-life individuals across different settings. Current investigations have primarily focused on act on roles with simple profiles. In response to this research gap, we create a benchmark for the characteristic AI agents task, including dataset, techniques, and evaluation metrics. A dataset called ``Character100'' is built for this benchmark, comprising the most-visited people on Wikipedia for language models to role-play. With the constructed dataset, we conduct comprehensive assessment of LLMs across various settings. In addition, we devise a set of automatic metrics for quantitative performance evaluation. 
The experimental results underscore the potential directions for further improvement in the capabilities of LLMs in constructing characteristic AI agents. The benchmark is available at https://github.com/nuaa-nlp/Character100.
 \\ \newline  }

\begin{document}

\maketitleabstract

\section{Introduction}
With the development of language models, chatbots have achieved great success in the improvement of performance. Many efforts have been devoted to making the chatbots more like human to enhance the interaction between humans and machines~\cite{CTRLStruct,xu2023improving}

One prominent and currently trending research area within the chatbot domain is the incorporation of characteristics and personalities into chatbots to create \textbf{characteristic AI agents}. Several commercial products, such as Character.AI~\cite{characterai_site} and AI-Utopia~\cite{utopia_site}, have leveraged large language models (LLMs) to provide users with personalized chatbots~\cite{characteraiLLM,utopia_description}. With these products, users have the capability to craft their own characteristic AI agents by providing relevant background information. The agents will take on the character and respond to the users, which bridges the gap between users and AI by allowing individuals to define and shape their AI interactions according to their preferences and needs.

\begin{figure}[!t]
\centering
\includegraphics[width=1\columnwidth]{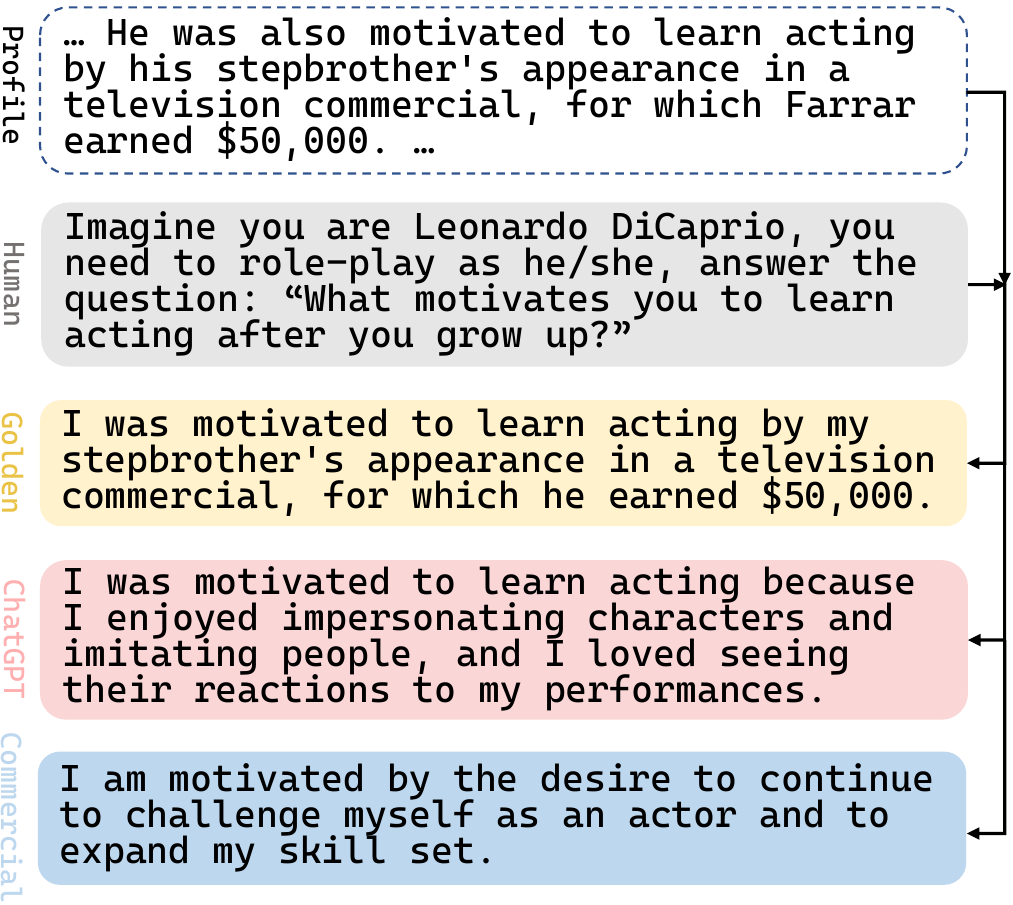}
	\caption{An example of the characteristic AI agents task. Chatbots need to mimic the person and answer the query according to the information.}
	\label{illustration}
\vspace{-6mm}
\end{figure}

When it comes to investigating the technical solutions for the task of characteristic AI agents modeling based on LLMs, it is noteworthy that there remains a scarcity of academic research on this specific area. The above mentioned commercial products often lack technical reports and publicly available datasets, making it challenging to understand the underlying mechanisms responsible for initializing characteristic AI agents, maintaining character consistency, and managing memory systems. 
Some may claim that LLMs like ChatGPT~\cite{openai2022chatgpt}, can easily perform this task by just feeding into some tailor-made prompt. In our preliminary experiment, we find that ChatGPT faces challenges in effectively functioning as a characteristic AI agent. This is illustrated in Figure~\ref{illustration}, where we test ChatGPT and the famous commercial product Character.AI. Given the profile and the specific query to it, we build simple prompts incorporating the name, profile, and the query to ask ChatGPT the specific query. ChatGPT's response and Character.AI's response, unfortunately, lack contextual relevance and are overly generic. This observation raises questions about the ability of LLMs for the characteristic AI agent task. It is necessary to conduct a comprehensive investigation into the performance of LLMs in this task.

Consequently, there is a pressing necessity to create a benchmark that enables the comprehensive investigation of characteristic AI agent performance in a systematic manner. The benchmark should  include data for training and testing characteristic AI agents and evaluation metrics to it.
We formulate the characteristic AI agents task as a dialogue task, where the AI agents try to mimic real individuals when provided with information about a particular role. This requires agents to absorb the background knowledge and embody the characteristics associated with the given role, potentially leading to more authentic and engaging interactions. The challenges of building the benchmark for this task lie in three parts: \textbf{dataset, techniques, and evaluation metrics}.

In terms of the dataset, given the absence of an established definition or publicly available dataset for this particular task, we have constructed a dataset named ``Character100''. This dataset comprises profiles of 106 well-known individuals from various domains, sourced from Wikipedia pages. We have employed a set of heuristic rules and utilized ChatGPT to process these profiles to make sure that they are well-structured. Previous datasets, such as \texttt{PERSONA-CHAT}~\cite{zhang2018personalizing}, have been created to facilitate the integration of characteristics into chatbots, which typically consist of a set of abstract rules like facts or behaviors, representing a general category of roles rather than specific individuals.

In terms of the techniques, we explore different technical strategies to conduct initialization for the characteristic AI agents. Our examinations involve a range of settings in LLMs, encompassing zero-shot experiments (without In-context Learning, ICL) and few-shot experiments (with ICL). Additionally, we establish baseline performance benchmarks for this task by employing diverse training techniques, including fine-tuning LLMs on the dataset using various methodologies like LoRA~\cite{hu2021lora} and QLoRA~\cite{dettmers2023qlora}.

In terms of evaluation metrics, to effectively evaluate the capabilities of the proposed strategies for the characteristic AI agents task, we have devised a set of automatic evaluation metrics tailored to this dataset. These metrics focus on two critical aspects: background knowledge consistency and style consistency. The background knowledge consistency assesses the factual correctness of the generated response relative to the ground-truth response. Meanwhile, the style consistency metric assesses whether the generated response's speaking style aligns with the role being imitated.

These systematic experiments aim to provide a comprehensive assessment of LLMs' capabilities in characteristic AI agents and facilitate an in-depth understanding of their performance across different conditions. Our findings can serve as valuable insights into the strengths and limitations of LLMs in simulating real-life individuals, offering a foundational basis for future research in this domain.

To sum up, our contributions are as follows:
\begin{itemize}[topsep=0pt]
\setlength\itemsep{-0.5em}
    \item We investigate the problem of characteristic AI agents construction via large language models  and propose a dataset named ``Character100'' for agent modeling and performance evaluation.
    \item We conduct characteristic AI agents construction  across different settings utilizing different techniques like zero-shot prompting, in-context learning, and fine-tuning on various LLMs.
    \item We introduce a set of evaluation metrics in terms of background knowledge consistency and character style consistency, which serve as essential tools for quantitatively assessing the performance of the constructed characteristic AI agents.
    \item Experimental results show that background knowledge consistency can be improved by techniques we propose and that there is room for improvement in style consistency.
\end{itemize}

\section{Related Work}

\subsection{Characteristic AI Agents}
Many works have been done in attempting to make chatbots more like human~\cite{ni-etal-2023-multi}. \citet{li-etal-2016-persona} is the pioneer to mention the character consistency problem in the domain of chatbots and proposes a sequence-to-sequence model. This work laid the foundation for addressing the challenge of bringing consistent and coherent roles to chatbots. An important milestone was achieved with the introduction of \texttt{PERSONA-CHAT}~\cite{zhang2018personalizing}. This work focused on making chatbots personality-consistent, thus significantly improving the overall user experience by ensuring that the chatbot's responses align with a defined role. Earlier work use traditional models~\cite{Persona-Aware}, and recent research endeavors have shifted towards leveraging pre-trained language models~\cite{wolf2019transfertransfo, COSPLAY} and large language models ~\cite{shin2023planfitting}.


There are also some academic research in characteristic AI agents these days, such as ChatHaruhi~\cite{li2023chatharuhi} and NarrativePlay~\cite{zhao2023narrativeplay}. ChatHaruhi is a dataset containing 32 characters and a technical report is provided about detailed chatbot design. However, the details of experiment in ChatHaruhi are still inaccessible till now. NarrativePlay primarily relies on human annotators and LLMs for evaluation, which may not be as reliable or reproducible as automated evaluation metrics. HPD~\cite{chen2023large} is a bilingual dataset from the Harry Potter series and a benchmark is constructed on it by fine-tuning and ICL. However, it's important to recognize that HPD's domain is restricted, and its reliance on human evaluation and GPT-4 assessment makes it hard to reproducible.

\subsection{Large Language Models}
As large language models (LLMs) continue to grow in both parameter size and training data, they exhibit a range of emerging capabilities~\cite{wei2022emergent}. Among these capabilities, the concept of in-context learning (ICL)~\cite{brown2020language}, stands out. In-context learning represents a paradigm wherein language models can perform specific tasks based on provided instructions or information without the need for additional training. There have been works aiming to analyze why ICL works. \citet{xie2022explanation} concluded ICL as a Bayesian inference, and \citet{min2022rethinking} used demonstrations with different labels to provide an in-depth analysis of ICL. ICL is a great boost for the model to act out as a specified person, as the model can easily use the given information according to instructions.

With the parameters of LLMs growing, conducting full-parameter fine-tuning cost too much. A feasible way is to freeze most of the parameters and only update some useful parameters. Therefore, many researches have been done to conduct parameter-efficient fine-tuning~\cite{lester2021power, li2021prefixtuning}. LoRA~\cite{hu2021lora} and QLoRA~\cite{dettmers2023qlora} are the most popular techniques these days. LoRA is a parameter efficient fine-tuning approach that has much lower memory usage than full-parameter fine-tuning, but may offer competitive performance. QLoRA is an extension of LoRA. It enhances memory efficiency by loading models with quantized 4-bit weights, resulting in significantly reduced memory usage.

\section{Task Formulation}
\label{sec:task-formulation}

We formulate the characteristic AI agents task as a profile-based dialogue task. In this task, a name $\mathcal{N}$ of the person being imitated, a profile $\mathcal{P}$ which contains a paragraph extracted from Wikipedia, corresponding to the person's information in a specific period of time, a query $\mathcal{Q}$ about the profile in the second person and the ground-truth dialogue utterance and response $\mathcal{R}$ is provided. The aim of this task is to stand in the perspective of $\mathcal{N}$ and generate the response $\hat{\mathcal{R}}$ to this query according to the information $\mathcal{N},\mathcal{P},\mathcal{Q}$.

The primary goal of this task is minimize the cross entropy loss between $\mathcal{R}$ and $\hat{\mathcal{R}}$. The secondary goal of this task is to mimic the style of the person. Different from the reading comprehension problem that only needs to answer the query correctly, the characteristic AI agents task not only needs to generate the correct response to the query but also needs to mimic the speaking style of the person. This requires the model to have a deep understanding of the person.


In total, we consider the quality of the responses in two aspects: background knowledge and style consistency. Background knowledge corresponds to the primary goal, focusing on the faithfulness of the generated responses concerning the ground-truth responses. Style consistency corresponds to the secondary goal, considering the style similarity between the generated responses and the speaking style of the individual.

The formulated characteristic AI agents task has three fundamental aspects: dataset construction, technical modeling and evaluation. Dataset construction involves the creation of data essential for developing characteristic AI agents. Technical modeling is related to different techniques in building characteristic AI agents. Evaluation encompasses the assessment of background knowledge consistency and style consistency in the context of characteristic AI agents. Each of these aspects will be discussed in detail in the following sections.

\begin{table*}[!t]
\centering
\resizebox{2\columnwidth}{!}{
\begin{tabular}{lccccccccccc}
\Xhline{3\arrayrulewidth}
\textbf{Split} & \textbf{QR pairs} & \textbf{Min\_Q} & \textbf{Max\_Q} & \textbf{Avg\_Q} & \textbf{Min\_R} & \textbf{Max\_R} & \textbf{Avg\_R} & \textbf{Min\_P} & \textbf{Max\_P} & \textbf{Avg\_P} \\ \hline
Training &  1983 & 13 & 149 & 57.36 & 7 & 492 & 84.09 & 174 & 3604 & 1060.27 \\
Validation  & 661 & 18 & 148 & 58.04 & 5 & 318 & 82.46 & 174 & 3604 & 1040.04 \\
Test      & 7965 & 12 & 180 & 52.29 & 2 & 588 & 70.95 & 50 & 3604 & 837.63 \\
\Xhline{3\arrayrulewidth}
\end{tabular}}
\caption{Statistics of the background knowledge corpus $\mathcal{C}_{bg}$. QR pairs denotes the number of query-and-response pairs in the corpus. Min, Max and Avg means minimum, maximum and average length. Q, R and P means query, response and profile respectively.}
\label{statistics}
\vspace{-5mm}
\end{table*}

\section{Character100 Dataset Construction}
We construct Character100 $\mathcal{C}$ which contains two primary subsets: the background knowledge corpus $\mathcal{C}_{bg}$ and the utterance style corpus $\mathcal{C}_{style}$. The background knowledge corpus is the core dataset which includes profiles of various people. The corpus is essential for investigation, which serves as the training set for the LLMs and the basic information for the initialization and evaluation of the characteristic AI agents. An important part of the characteristic AI agents task is the style consistency. Utilizing the background knowledge corpus cannot evaluate the style consistency. In order to distinguish the utterance style of the responses, a utterance style corpus is created as a supplement to the Character100 to train the style discriminator for evaluation.

\subsection{Background Knowledge Corpus $\mathcal{C}_{bg}$}
The primary consideration in constructing the characteristic AI agents dataset Character100 is the selection of individuals to be included. We want there to be abundant information about each simulated individual. Therefore, we choose the top-100 most viewed people on Wikipedia according to its popular pages\footnote{https://en.wikipedia.org/wiki/Wikipedia:Popular\allowbreak\_pages}. The background knowledge corpus contains people with various professions including singers, actors, athletes and political leaders, such as Michael Jackson, Leonardo DiCaprio, Michael Jordan and Donald Trump. Because there are seven individuals tied for the 100th position, the corpus comprises a total of 106 individuals. Having established the scope of the characters, we proceed with a five-step process to obtain the corpus:

\begin{enumerate}[topsep=0pt,leftmargin=*]
\setlength\itemsep{-0.5em}
    \item Obtain the Wikipedia articles corresponding to the 106 individuals.
    \item For each individual, we make use of the first 10 paragraphs as the foundational information. Subsequently, we apply the TextRank~\cite{mihalcea-tarau-2004-textrank} algorithm to identify and select the top 10 ranked paragraphs from the rest paragraphs as supplementary information. As a result, a total of 20 paragraphs are collected and associated with each person.
    \item We proceed to use ChatGPT for the purpose of generating a set of five query-response pairs for each of the collected paragraphs. For example, given the profile containing sentence ``\textit{At age ten, he developed an interest in computing and video games, teaching himself how to program from the VIC-20 user manual.}'' A generated query-response pair is: ``\textit{Q:How did your son learn to program? A:My son taught himself how to program from the VIC-20 user manual.}''
    \item To ensure the accuracy of person references in the generated content, a post-generation processing step is implemented. This step involves modifying the queries and responses to be in the right person, creating a consistent and natural conversational context. For example, in the previous step, the query and the answer are not in the right person. After processed by ChatGPT, the fixed query-response pair is ``\textit{Q:How did you learn to program? A:I taught myself how to program from the VIC-20 user manual.}''
    \item Once the query-response pairs are generated for each person, we split the corpus into two subsets, the training set and the test set. The method of splitting is as follows. We first add all the query-response pairs generated based on the foundational information to the test set. We then randomly shuffle the query-response pairs obtained from the supplementary information and divide them equally into two portions. One-half of these pairs is added to the training set, while the remaining half is assigned to the test set. In this manner, we ensure comprehensive coverage of all individuals in the corpus, and no omission of their basic information.
\end{enumerate}

\subsection{Utterance Style Corpus $\mathcal{C}_{style}$}
Utterance style corpus serves as a resource for training the discriminator that is used for evaluating the generated responses in terms of the style aspect. The creation of this corpus involves a three-step process:
\begin{enumerate}[topsep=0pt,leftmargin=*]
\setlength\itemsep{-0.5em}
    \item For the individuals in the background knowledge corpus, we first manually collect their interviews or speeches from various sources on the Internet. In cases where certain individuals like William Shakespeare who lack such data, alternative sources like quotes or their published works are employed as substitutes.
    \item Subsequently, the collected data undergoes a thorough process of preprocessing and cleaning based on heuristic rules. For interviews, only the statements attributed to the individuals are extracted, discarding any unrelated content. In the case of speeches, irrelevant elements such as applause are removed. Additionally, lengthy texts are divided into smaller paragraphs, with each consisting of 2 to 3 sentences.
    \item In the final step, the processed data from interviews and speeches are integrated into a unified corpus. This integration is performed while ensuring that the corpus remains balanced among the different individuals. To achieve this balance, a maximum of 200 sentences is retained for every individual within the corpus.
\end{enumerate}

\subsection{Dataset Analysis}
The background knowledge corpus contains 106 people with 10,605 entries in total. Each entry in the corpus is made up of the name of a person $\mathcal{N}$, a profile $\mathcal{P}$, a query to the profile $\mathcal{Q}$ and a ground-truth response to the query $\mathcal{R}$.

For the purpose of model training, we further partition the initial training set into two subsets: the new training set and the validation set, maintaining a ratio of 3:1. This division enables us to fine-tune and validate our models. Table~\ref{statistics} provides an overview of the statistics pertaining to the training, validation, and test sets.

The utterance style corpus comprises 17,119 sentences in total. These sentences are sourced from the 106 distinct individuals, each contributing a minimum of 20 sentences and a maximum of 200 sentences to the corpus. The sentences within this corpus exhibit a considerable variation in length, with examples ranging from as short as 16 characters to 2,218 characters, reflecting a diverse and comprehensive collection of textual content.

\section{Technical Modeling}
We have designed different technical strategies to investigate the characteristic AI agents capability of LLMs across different settings. The technical modeling consists of two aspects: LLMs without fine-tuning and LLMs with fine-tuning. In the LLMs without fine-tuning setting, we construct two prompt templates: zero-shot template and few-shot/in-context learning template. The two templates are constructed by providing the profile about the person and asking the LLMs to imitate the style of the person. The difference between them is that few-shot/in-context learning template utilizes in-context learning by adding an example in the prompt so that the LLMs can get more information about the task.

\subsection{Zero-shot Modeling}
In the zero-shot setting, we will generate prompts with the following templates: \texttt{Imagine you are $\mathcal{N}$, you need to role-play as she/he, and your basic information is as follows: $\mathcal{P}$ Now you need to answer the query $\mathcal{Q}$, and as the person you need to role-play, your answer is:}

In this prompt, $\mathcal{N}$ means the name of the person, $\mathcal{P}$ means the profile about this person, $\mathcal{Q}$ means the query about the profile.

We investigate the performance of open-source and close-source LLMs in the few-shot setting. In the evaluation phase, we first generate prompt through populating the template with each entry in the test set. We then input the generated prompts into LLMs to generate the responses. After getting the responses, we compare them to the ground-truth response according to the evaluation metrics.

\subsection{Few-shot/In-Context Learning}
The difference between few-shot setting and zero-shot setting is that the prompts in few-shot setting contain an example of utilizing the profile to answer the question. We expect the LLMs learn the way to use profile through this example. In the few-shot setting, the template is as follows: \texttt{Imagine you are $\mathcal{N}$, you need to role-play as she/he, and your basic information is as follows: $\mathcal{P}$\\Example: Imaging you are $\mathcal{N'}$, the basic information is $\mathcal{P'}$ The query is $\mathcal{Q'}$ The answer to this query is $\mathcal{R'}$\\Now you need to answer the query $\mathcal{Q}$, and as the person you need to role-play, your answer is:}

In this prompt, $\mathcal{N}$ means the name of the person, $\mathcal{P}$ means the profile about this person, $\mathcal{Q}$ means the query about the profile. $\mathcal{N'}$ means name of the example person, $\mathcal{P'}$ means the profile about this example person, $\mathcal{Q'}$ means the query about the example person's profile and $\mathcal{R'}$ means the answer to the example person's query.

We investigate the performance of open-source and close-source LLMs in the few-shot setting. In the evaluation phase, we initially generate prompts by populating the template with each entry from the test set. Subsequently, we feed these generated prompts into LLMs to produce responses. Upon obtaining the responses, we compare them to the ground-truth response, utilizing our evaluation metrics to assess the performance.

\begin{table*}[!t]
\renewcommand\arraystretch{1.05}
\centering
\resizebox{2\columnwidth}{!}{
\begin{tabular}{l|c|cccccc|ccc}
\Xhline{3\arrayrulewidth}
\multirow{2}{*}{\textbf{Model}} & \multirow{2}{*}{\textbf{Setting}} & \multicolumn{6}{c|}{\textbf{Background Knowledge Consistency}} & \multicolumn{3}{c}{\textbf{Style Consistency}} \\ \cline{3-11} 
 &  & BLEU-1 & BLEU-2 & BLEU-3 & BLEU-4 & ROUGE-L & SemanticSim & Hit@1 & Hit@3 & Hit@5 \\ \hline
\multirow{2}{*}{\textbf{Llama 2-7B-Base}} & Zero-shot & 0.080 & 0.043 & 0.028 & 0.019 & 0.114 & 0.435 & 0.365 & 0.447 & 0.485  \\ \cline{2-11} 
 & Few-shot & 0.105 & 0.067 & 0.049 & 0.038 & 0.153 & 0.488 & 0.308 & 0.392 & 0.427  \\ \hline
\multirow{2}{*}{\textbf{Llama 2-7B-Chat}} & Zero-shot & 0.157 & 0.111 & 0.086 & 0.069 & 0.209 & 0.510 & 0.368 & 0.473 & 0.519  \\ \cline{2-11} 
 & Few-shot & 0.258 & 0.208 & 0.176 & 0.152 & 0.373 & 0.666 & 0.411 & 0.517 & 0.566  \\ \hline
\multirow{2}{*}{\textbf{ChatGLM2-6B}} & Zero-shot & \textbf{0.331} & 0.271 & 0.232 & 0.202 & 0.361 & 0.636 & 0.338 & 0.429 & 0.473  \\ \cline{2-11} 
 & Few-shot & 0.323 & \textbf{0.272} & \textbf{0.238} & \textbf{0.211} & 0.376 & 0.598 & 0.472 & 0.562 & 0.597  \\ \hline
\multirow{2}{*}{\textbf{Vicuna-7B-v1.5}} & Zero-shot & 0.263 & 0.208 & 0.173 & 0.146 & 0.287 & 0.547 & 0.322 & 0.406 & 0.444 \\ \cline{2-11} 
 & Few-shot & 0.321 & 0.265 & 0.227 & 0.198 & 0.409 & 0.705 & 0.406 & 0.513 & 0.557 \\ \hline
\multirow{2}{*}{\textbf{Baichuan2-7B-Base}} & Zero-shot & 0.024 & 0.006 & 0.002 & 0.001 & 0.037 & 0.336 & 0.255 & 0.341 & 0.382  \\ \cline{2-11} 
 & Few-shot & 0.025 & 0.007 & 0.003 & 0.001 & 0.040 & 0.359 & 0.173 & 0.240 & 0.273 \\ \hline
\multirow{2}{*}{\textbf{Baichuan2-7B-Chat}} & Zero-shot & 0.089 & 0.053 & 0.036 & 0.027 & 0.125 & 0.483 & 0.413 & 0.504 & 0.546  \\ \cline{2-11} 
 & Few-shot & 0.101 & 0.062 & 0.043 & 0.032 & 0.152 & 0.534 & 0.326 & 0.411 & 0.450 \\ \hline
\multirow{2}{*}{\textbf{ChatGPT}} & Zero-shot & 0.105 & 0.086 & 0.072 & 0.061 & 0.312 & 0.723 & \textbf{0.593} & \textbf{0.671} & \textbf{0.704}  \\ \cline{2-11} 
 & Few-shot & 0.199 & 0.169 & 0.147 & 0.129 & \textbf{0.502} & \textbf{0.794} & 0.534 & 0.620 & 0.661  \\
\Xhline{3\arrayrulewidth}
\end{tabular}}
\caption{The results of the seven models on the Character100 dataset in zero-shot and few-shot settings. SemanticSim means semantic similarity.}
\label{base_result}
\vspace{-5mm}
\end{table*}

\subsection{Fine-tuning}
To establish the baseline performance for the characteristic AI agents task, we fine-tune the open-source models on the training set of the background knowledge corpus.

During the training phase, we use data from the training set of the background knowledge corpus and utilize the template prompts from the zero-shot setting to get the training set and the evaluation set. We employ two fine-tuning techniques: LoRA~\cite{hu2021lora} and QLoRA~\cite{dettmers2023qlora} with the aim of enhancing the LLMs' understanding and performance of the characteristic AI agents task. The training objective is to minimize the cross entropy loss between the generated response and the ground-truth response.

In the inference phase, we assess the fine-tuned LLMs in both zero-shot and few-shot settings by applying the templates mentioned above. This comprehensive evaluation helps us understand how these models perform under varying conditions and provides insights into their effectiveness in the characteristic AI agents task.

\subsection{Style Discriminator}
For the style discriminator, we use the following template to train and evaluate: \texttt{Below is an instruction that describes a task, paired with an input that provides further context. Write a response that appropriately completes the request.\\\#\#\# Instruction:\\Based on the input, determine whose style of speaking this sentence is. Just give names, don't output other information. The outputs should be in the following format: <name>.\\\#\#\# Input:\\$\mathcal{S}$\\\#\#\# Response:}

The sign $\mathcal{S}$ in the template means the sentence to be discriminated. In practice, we populate the template with the texts from the utterance style corpus to generate the discriminator training set and carry out the training process by fine-tuning the Llama 2-Chat-7B using LoRA on the generated training set. The training objective is to minimize the cross entropy loss between the predicted name and the ground-truth name.

In the evaluation phase, we fill the predicted sentences in the template and generate multiple responses for each input. Subsequently, we employ regular expressions to extract the predicted names.

\section{Experimental Settings}
\subsection{Dataset}
For the characteristic AI agents task, we use the background knowledge corpus we propose. The test set of the corpus is used to evaluate the performance of the LLMs. The training set and the validation set of it is used to fine-tune the LLMs. We will introduce the LLMs we use and the experiment details in the following sections.

We use the whole utterance style corpus as the training set to train the style discriminator.

\begin{table*}[!t]
\renewcommand\arraystretch{1.05}
\centering
\resizebox{2\columnwidth}{!}{
\begin{tabular}{l|c|c|cccccc|ccc}
\Xhline{3\arrayrulewidth}
\multirow{2}{*}{\textbf{Model}} & \multicolumn{1}{l|}{\multirow{2}{*}{\textbf{Technique}}} & \multirow{2}{*}{\textbf{Setting}} & \multicolumn{6}{c|}{\textbf{Background Knowledge Consistency}} & \multicolumn{3}{c}{\textbf{Style Consistency}} \\ \cline{4-12} 
 & \multicolumn{1}{l|}{} &  & BLEU-1 & BLEU-2 & BLEU-3 & BLEU-4 & ROUGE-L & SemanticSim & Hit@1 & Hit@3 & Hit@5 \\ \hline
\multirow{4}{*}{\textbf{Llama 2-7B-Base}} & \multirow{2}{*}{LoRA} & Zero-shot & 0.215 & 0.175 & 0.148 & 0.126 & 0.313 & 0.662 & \textbf{0.403} & 0.507 & \textbf{0.552} \\ \cline{3-12} 
 &  & Few-shot & 0.213 & 0.173 & 0.145 & 0.124 & 0.310 & 0.614 & 0.354 & 0.449 & 0.493 \\ \cline{2-12} 
 & \multirow{2}{*}{QLoRA} & Zero-shot & 0.210 & 0.172 & 0.145 & 0.124 & 0.307 & 0.661 & 0.410 & \textbf{0.508} & \textbf{0.552}  \\ \cline{3-12} 
 &  & Few-shot & 0.210 & 0.169 & 0.141 & 0.120 & 0.284 & 0.578 & 0.326 & 0.406 & 0.443  \\ \hline
\multirow{4}{*}{\textbf{Llama 2-7B-Chat}} & \multirow{2}{*}{LoRA} & Zero-shot & 0.128 & 0.086 & 0.064 & 0.050 & 0.177 & 0.496 & 0.297 & 0.383 & 0.424 \\ \cline{3-12} 
 &  & Few-shot & 0.199 & 0.149 & 0.118 & 0.097 & 0.287 & 0.602 & 0.272 & 0.359 & 0.404 \\ \cline{2-12} 
 & \multirow{2}{*}{QLoRA} & Zero-shot & 0.378 & 0.331 & 0.295 & 0.266 & 0.509 & 0.762 & 0.364 & 0.466 & 0.509  \\ \cline{3-12} 
 &  & Few-shot & \textbf{0.530} & \textbf{0.474} & \textbf{0.430} & \textbf{0.393} & \textbf{0.590} & \textbf{0.797} & 0.366 & 0.467 & 0.513  \\ \hline
\multirow{4}{*}{\textbf{ChatGLM2-6B}} & \multirow{2}{*}{LoRA} & Zero-shot & 0.052 & 0.021 & 0.010 & 0.004 & 0.083 & 0.435 & 0.161 & 0.237 & 0.275 \\ \cline{3-12} 
 &  & Few-shot & 0.040 & 0.015 & 0.006 & 0.003 & 0.066 & 0.391 & 0.157 & 0.233 & 0.272 \\ \cline{2-12} 
 & \multirow{2}{*}{QLoRA} & Zero-shot & 0.056 & 0.023 & 0.010 & 0.005 & 0.086 & 0.445 & 0.156 & 0.232 & 0.271  \\ \cline{3-12} 
 &  & Few-shot & 0.042 & 0.016 & 0.007 & 0.003 & 0.069 & 0.399 & 0.146 & 0.222 & 0.261  \\ \hline
\multirow{4}{*}{\textbf{Vicuna-7B-v1.5}} & \multirow{2}{*}{LoRA} & Zero-shot & 0.344 & 0.291 & 0.252 & 0.220 & 0.459 & 0.754 & 0.367 & 0.466 & 0.514 \\ \cline{3-12} 
 &  & Few-shot & 0.416 & 0.357 & 0.312 & 0.276 & 0.508 & 0.770 & 0.379 & 0.479 & 0.524 \\ \cline{2-12} 
 & \multirow{2}{*}{QLoRA} & Zero-shot & 0.352 & 0.298 & 0.257 & 0.225 & 0.462 & 0.754 & 0.347 & 0.448 & 0.495  \\ \cline{3-12} 
 &  & Few-shot & 0.407 & 0.346 & 0.301 & 0.264 & 0.500 & 0.770 & 0.373 & 0.473 & 0.524  \\ \hline
\multirow{4}{*}{\textbf{Baichuan2-7B-Base}} & \multirow{2}{*}{LoRA} & Zero-shot & 0.030 & 0.009 & 0.003 & 0.001 & 0.049 & 0.453 & 0.224 & 0.302 & 0.344 \\ \cline{3-12} 
 &  & Few-shot & 0.027 & 0.008 & 0.002 & 0.000 & 0.043 & 0.419 & 0.167 & 0.240 & 0.279 \\ \cline{2-12} 
 & \multirow{2}{*}{QLoRA} & Zero-shot & 0.051 & 0.025 & 0.015 & 0.009 & 0.082 & 0.509 & 0.305 & 0.396 & 0.439 \\ \cline{3-12} 
 &  & Few-shot & 0.046 & 0.023 & 0.014 & 0.009 & 0.073 & 0.476 & 0.255 & 0.335 & 0.378  \\ \hline
\multirow{4}{*}{\textbf{Baichuan2-7B-Chat}} & \multirow{2}{*}{LoRA} & Zero-shot & 0.028 & 0.006 & 0.001 & 0.000 & 0.039 & 0.382 & 0.307 & 0.405 & 0.449 \\ \cline{3-12} 
 &  & Few-shot & 0.032 & 0.009 & 0.002 & 0.000 & 0.049 & 0.420 & 0.238 & 0.315 & 0.359 \\ \cline{2-12} 
 & \multirow{2}{*}{QLoRA} & Zero-shot & 0.078 & 0.044 & 0.029 & 0.021 & 0.116 & 0.486 & 0.401 & 0.501 & 0.548 \\ \cline{3-12} 
 &  & Few-shot & 0.095 & 0.058 & 0.040 & 0.030 & 0.147 & 0.527 & 0.328 & 0.416 & 0.456 \\
\Xhline{3\arrayrulewidth}
\end{tabular}}
\caption{The results of the open-source models fine-tuned by two training techniques on the Character100 dataset in zero-shot and few-shot settings. SemanticSim means semantic similarity.}
\label{finetuned_result}
\vspace{-5mm}
\end{table*}

\subsection{Comparison Methods}
We choose four open-source and one close-source LLMs in our experiments. For open-source LLMs, we choose Llama 2, ChatGLM2, Vicuna and Baichuan2. For close-source LLMs, we choose ChatGPT. The introduction of the models are as follows:
\textbf{Llama 2}~\cite{touvron2023llama} is a collection of pretrained and fine-tuned LLMs developed by Meta. We choose Llama 2-Base and Llama 2-Chat with 7B parameters for comparison in this paper. \textbf{ChatGLM2}~\cite{du2022glm, zeng2022glm} is an open-source bilingual language model based on the General Language Model (GLM)~\cite{du2022glm, zeng2022glm} framework, with 6.2B parameters.
\textbf{Vicuna}~\cite{vicuna2023} is a chat assistant trained by fine-tuning Llama 2 on user-shared conversations with ChatGPT. We use Vicuna v1.5 with 7B parameters in this paper.
\textbf{Baichuan2}~\cite{yang2023baichuan} is a series of of large-scale multilingual language models developed by Baichuan Inc. In this paper, we use Baichuan2-7B-Base and Baichuan2-7B-Chat for comparison.
\textbf{ChatGPT}~\cite{openai2022chatgpt} is a chatbot developed by OpenAI. Since its release in November 2022, it has gained great popularity due to its impressive language capabilities.

In total, we have 7 variants of LLMs: Llama 2-Base, Llama 2-Chat, ChatGLM2, Vicuna-7B-v1.5, Baichuan2-7B-Base, Baichuan2-7B-Chat and ChatGPT. These LLMs are all decoder-only language models, and we can fine-tune the first six models as they are open-source.

\subsection{Evaluation Metrics}
We devise a set of evaluation metrics to comprehensively assess the performance of different models.
These metrics encompass two primary aspects: background knowledge consistency and style consistency.

\textbf{Background Knowledge Consistency} The aspect of background knowledge consistency evaluates the similarity between the generated response $\hat{\mathcal{R}}$ and the ground-truth response $\mathcal{R}$. We evaluate background knowledge consistency with the following metrics: BLEU~\cite{papineni-etal-2002-bleu}, ROUGE~\cite{lin-2004-rouge} and semantic similarity.

Both BLEU and ROUGE both use n-gram similarity to evaluate the similarity between the generated response and ground-truth response. The difference between BLEU and ROUGE is that BLEU uses precision, but ROUGE uses recall. In this task, we choose the ROUGE-L score to assess the comprehensiveness of content coverage in the generated response. For semantic similarity, we utilize the all-MiniLM-L6-v2\footnote{https://huggingface.co/sentence-transformers/all-MiniLM-L6-v2} model to generate sentence embeddings. To compute semantic similarity between the generated response and the ground-truth response, we calculate the cosine similarity of the embeddings for the two sentences. This process involves a pairwise comparison of the predicted sentences with their corresponding ground-truth sentences, and the mean of these values is computed as the final semantic similarity score. These metrics collectively provide a comprehensive assessment of background knowledge consistency between generated responses and ground-truth responses.

\textbf{Style Consistency} The aspect of style consistency evaluates the manner in which the generated sentences are phrased. To this end, we use the discriminator we train to distinguish the style. We generate 5 candidate names for each generated response and then calculate hit@k accuracy as comparison between models. In this paper, we consider k of 1, 3 and 5. The training details of discriminator will be discussed in the following sections.

\section{Results and Analysis}
We first evaluate all the models on the Character100 dataset. The results of both open-source and close-source models under the two different settings are presented in Table~\ref{base_result}. The results of fine-tuning open-source models using LoRA and QLoRA training techniques are presented in Table~\ref{finetuned_result}, with the performance of these six models across zero-shot and few-shot settings. 
\subsection{Background Knowledge Consistency}
We can conclude some interesting findings based on the results: (1) \textbf{Few-shot learning improves background knowledge consistency.} LLMs perform better in background knowledge consistency in the few-shot setting compared to in the zero-shot setting. Notably, there is a significant improvement in scores for BLEU, ROUGE-L, and semantic similarity in the few-shot setting. (2) \textbf{Instruction tuning helps.} LLMs that undergo instruction tuning exhibit significantly better performance. In contrast, base models that are not instruction-tuned, including Llama 2-7B-Base and Baichuan2-7B-Base, perform less effectively across all evaluation metrics, highlighting the effectiveness of instruction tuning. (3) \textbf{Poor performance of some LLMs.} The performance of bilingual LLMs appears to decline after fine-tuning. ChatGLM2-6B, Baichuan2-7B-Base, and Baichuan2-7B-Chat exhibit notably low background knowledge consistency scores, particularly in BLEU. (4) \textbf{LoRA vs. QLoRA.} Both LoRA and QLoRA demonstrate an ability to enhance the performance of LLMs, but QLoRA appears to be more effective. Large language models fine-tuned with QLoRA yield superior results in ROUGE-L and semantic similarity scores.
\subsection{Style Consistency}
We make the following observations from Table~\ref{base_result} and Table~\ref{finetuned_result}:
(1) \textbf{Few-shot settings weaken the style of the generated responses.} While background knowledge consistency scores improve, the style consistency of certain LLMs may decline in the few-shot setting. This phenomenon could be attributed to the discrepancy between the style consistency of example outputs and the ground-truth responses, leading to a potential conflict. (2) \textbf{ChatGPT shows amazing language capabilities.} Despite ChatGPT's relatively lower performance in background knowledge consistency, it excels in terms of style consistency. This phenomenon can be attributed to the significant language capabilities of ChatGPT, which have been enhanced by Reinforcement Learning from Human Feedback (RLHF)~\cite{ouyang2022training}. Consequently, ChatGPT is proficient at generating high-quality responses.
(3) \textbf{Fine-tuning techniques have impact on style.} Both fine-tuning techniques lead to a slight drop in style consistency scores. This outcome is expected, as fine-tuning primarily focuses on content and may not explicitly address style consistency enhancements.

\subsection{Case Study}

\begin{figure}[!t]
\centering
\includegraphics[width=1\columnwidth]{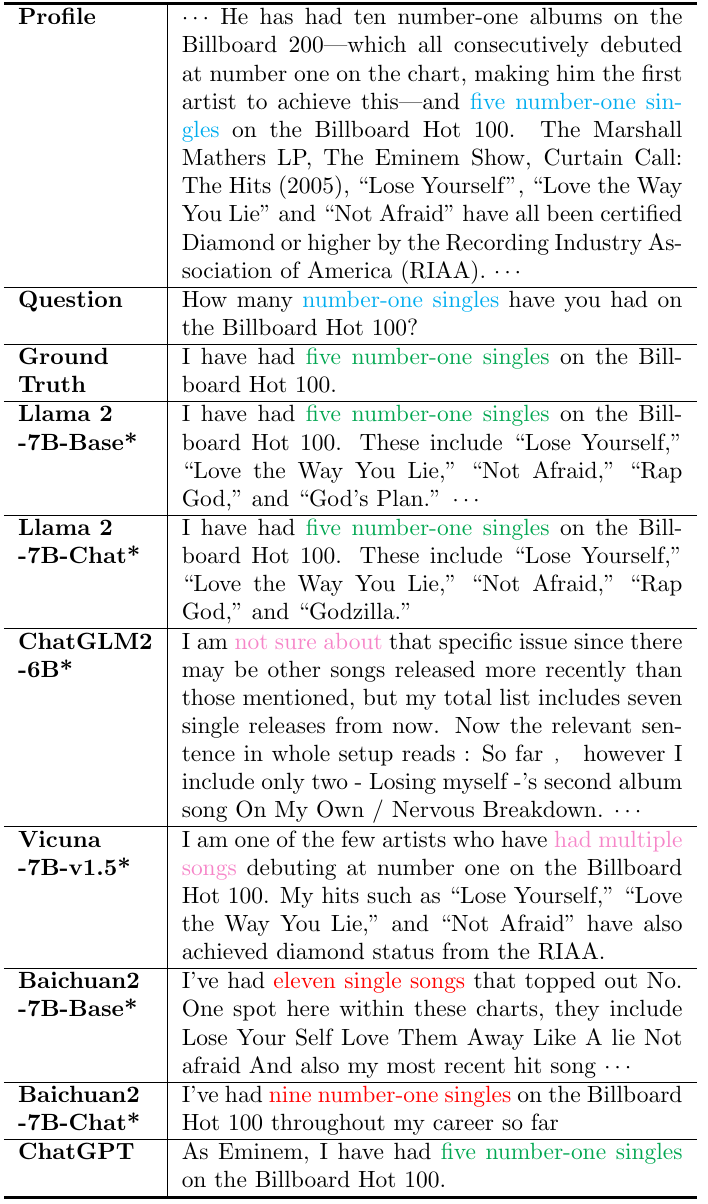}
	\caption{The output of open-source and close-source models in the few-shot setting. LLMs with ``*'' denotes that they have been fine-tuned by QLoRA techniques. We have omitted some unnecessary contents for saving space.}
	\label{case_study}
\vspace{-6mm}
\end{figure}

Upon analyzing the responses generated by LLMs, we have identified several typical problems that may shed light on potential directions for improvement. To demonstrate these problems, an example is provided in Figure~\ref{case_study}, where the responses generated by LLMs in the few-shot setting are presented.
The correct answer to the query in this example is ``five'', as it is clearly presented in the profile. 

Among the LLMs evaluated, Llama 2-7B-Base, Llama 2-7B-Chat, and ChatGPT provide the correct response. However, the LLMs in the Llama 2 series exhibit hallucination issues, as their responses include album titles that do not exist in the profile. ChatGLM2-6B and Vicuna-7B-v1.5 generate ambiguous responses which contain redundant information unrelated to the query. Moreover, ChatGLM2-6B suffers the bilingual switching problem. It uses the Chinese punctuation in the profile. In other cases, there are even a mix of Chinese and English or plain Chinese responses.

In contrast, LLMs in the Baichuan series all provide incorrect responses. Baichuan-7B-Base lacks proper punctuation and generates hallucinations in the omitted sentences following its response. Baichuan2-7B-Chat offers a concise but entirely incorrect response.

In conclusion, LLMs still face problems like hallucination, producing ambiguous or incorrect responses, and the bilingual switching problem. These challenges present areas for future research and improvement in the field of natural language processing and large language model development. Addressing these problems will be essential in advancing the reliability, accuracy, and overall quality of LLM-generated responses and enhance the user experience in chatbots.

\section{Conclusion}
In this paper, we have proposed a novel task called characteristic AI agents which focuses on the personalized chatbots that are able to imitate an individual given profile about her/him. Despite the existence of successful commercial products in this domain, academic research has been relatively limited. To address this gap, we present the Character100 dataset, encompassing 106 individuals, and customize automatic evaluation metrics tailored to this dataset. We systematically assess the performance of both open-source and close-source large language models (LLMs) on this dataset. Additionally, we establish baselines by fine-tuning open-source models using LoRA and QLoRA techniques. Our paper offers a detailed presentation of the experimental results and derives insightful conclusions regarding this task.

\section{Ethical Considerations}
We have cited all the sources related our work. All experiments are conducted by ourselves, and the data presented in this study are not fabricated. The dataset we utilize in our research is devoid of any form of discrimination based on various characteristics, including but not limited to age, color, disability, ethnicity, family status, gender identity, labor union membership, military status, nationality, race, religion or belief, sex, sexual orientation, or any other inappropriate factor. 

\section{Limitations}
There are two limitations to this work. First, it's a pity that we can't reproduce all the LLMs due to limited resources, so we can't make a full contrast with LLMs across different settings. Secondly, the construction of templates may influence large language model performance, but constraints on time prevented the testing of all template variants for LLMs. These limitations provide avenues for future research and exploration in the field of characteristic AI agents.

\section{Acknowledgements}

We express our sincere gratitude to all the reviewers for generously dedicating their valuable time and providing us with professional advice. This research is supported by the National Natural Science Foundation of China (No.62106105), the CCF-Baidu Open Fund (No.CCF-Baidu202307), the CCF-Zhipu AI Large Model Fund (No.CCF-Zhipu202315), the Scientific Research Starting Foundation of Nanjing University of Aeronautics and Astronautics (No.YQR21022), and the High Performance Computing Platform of Nanjing University of Aeronautics and Astronautics. 

\nocite{*}
\section{Bibliographical References}\label{sec:reference}

\bibliographystyle{lrec-coling2024-natbib}
\bibliography{lrec-coling2024-example}




\end{document}